\documentclass[conference]{IEEEtran}
\IEEEoverridecommandlockouts
\usepackage{cite}
\usepackage{amsmath,amssymb,amsfonts}
\usepackage{algorithmic}
\usepackage{graphicx}
\usepackage{textcomp}
\usepackage{xcolor}
\def\BibTeX{{\rm B\kern-.05em{\sc i\kern-.025em b}\kern-.08em
    T\kern-.1667em\lower.7ex\hbox{E}\kern-.125emX}}

\usepackage{amsmath}
\usepackage[ruled,vlined, linesnumbered]{algorithm2e}
\usepackage{pgf}
\usepackage{tikz}
\usetikzlibrary{arrows,automata}
\usepackage[font=small]{subcaption}
\usepackage{multirow}
\usepackage{booktabs}
\usepackage{arydshln}
\usepackage{optidef}

\usepackage{verbatim}
\usetikzlibrary{arrows,decorations.pathmorphing,backgrounds,positioning,fit,matrix}

\begin{document}

\IEEEoverridecommandlockouts

\IEEEpubid{\makebox[\columnwidth]{978-1-7281-8763-1/20/\$31.00~\copyright{}2020 IEEE \hfill} \hspace{\columnsep}\makebox[\columnwidth]{ }}

\title{Path  Planning  Followed  by  Kinodynamic Smoothing  for  Multirotor  Aerial  Vehicles (MAVs)\\
}
\newcommand{\linebreakand}{%
  \end{@IEEEauthorhalign}
  \hfill\mbox{}\par
  \mbox{}\hfill\begin{@IEEEauthorhalign}
}
\makeatother
\author{
  \IEEEauthorblockN{Geesara Kulathunga, Dmitry Devitt, Roman Fedorenko,  Sergei Savin and Alexandr Klimchik}
  \IEEEauthorblockA{
   \textit{Center for Technologies in Robotics and Mechatronics Components, Innopolis University, Russia} \\
    ggeesara@gmail.com,  d.dmitry@innopolis.ru, r.fedorenko@innopolis.ru, s.savin@innopolis.ru, a.klimchik@innopolis.ru}
    }

\maketitle

\begin{abstract}
We explore path planning followed by kinodynamic smoothing while ensuring the vehicle dynamics feasibility for MAVs. We have chosen a geometrically based motion planning technique \textquotedblleft RRT*\textquotedblright\; for this purpose. In the proposed technique, we modified original RRT* introducing an adaptive search space and a steering function which help to increase the consistency of the planner. Moreover, we propose multiple RRT* which generates a set of desired paths, provided that the optimal path is selected among them. Then, apply kinodynamic smoothing, which will result in dynamically feasible as well as obstacle-free path. Thereafter, a b spline-based trajectory is generated to maneuver vehicle autonomously in unknown environments. Finally, we have tested the proposed technique in various simulated environments. 
\end{abstract}


\section{Introduction}

 With the recent research advances in microcontroller technology and sensors capabilities, a new era has begun for MAVs. MAVs have been engaging with the plenty of applications including delivery, farming and cinematography in the recent past. Motion planning is one of the challenging tasks in almost all preceding scenarios. Subsequently, geometric based motion planning is a well-matured technique although no differential constraints (i.e., vehicle dynamics) are considered. Conversely, kinodynamics motion planning is one of the ways to ensure vehicle dynamics feasibility and fulfilling all the given constraints. When considering real-time motion planning, geometric based path planning followed by path parameterization is a well-adapted technique. But in this approach, path parameterization may fail despite selecting geometrically shortest path or the optimal selection of path finding algorithm. 
 

\begin{figure}[ht!]
 \begin{subfigure}[b]{1\linewidth}
    \centering
    \includegraphics[width=0.6\linewidth]{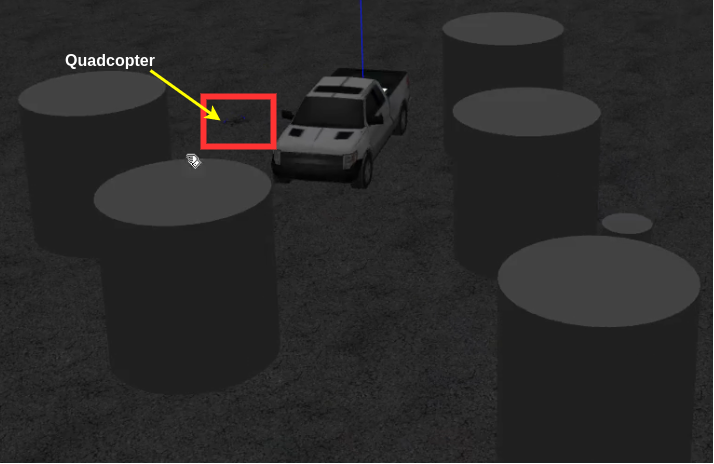}
    \caption{A sample environment that is used for testing the proposed replanner}
    \label{fig:simulated_environment}
  \end{subfigure}%
  \hfill
  \begin{subfigure}[b]{1\linewidth}
    \centering
    \includegraphics[width=0.6\linewidth]{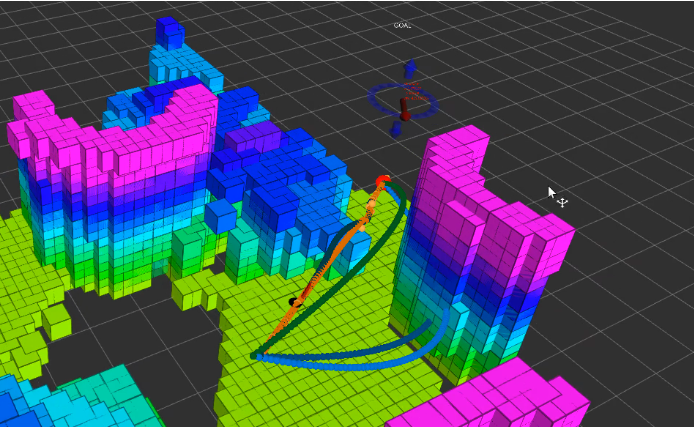}
    \caption{When the euclidean distance between start and goal pose less than the prediction horizon of the RRT*, it will generate a full path between start and goal pose}
    \label{fig:full_path}
  \end{subfigure}%
  \hfill
  \begin{subfigure}[b]{1\linewidth}
    \centering
    \includegraphics[width=0.6\linewidth, ]{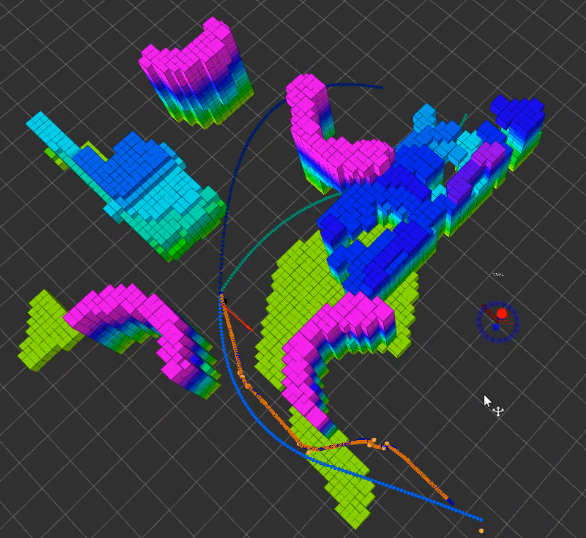}
    \caption{When the euclidean distance between start and goal pose higher than the horizon of the RRT*, it will generate a path in between start and horizon as shown here. The nominal path is re-projected whenever close obstacles are identified}
     \label{fig:horizon}
  \end{subfigure}
  \caption[Obs Point 1]{Improved RRT* generates N (i.e., 4) number of paths and select the optimal path. Afterwards, iLQR smoothing is applied on the selected path depicts in orange color. The other paths are shown in blue, dark blue and green respectively whereas small red color sphere is the goal pose}
  \label{f:simulated_results}
\end{figure}


In this paper, we explore kinodynamic smoothing followed by trajectory generation to keep the dynamic feasibility while reducing the execution time compared to kinodynamic motion planning. We selected a sampling-based technique,  RRT*~\cite{karaman2011sampling} comes under geometric based planning that guarantees asymptotic optimality. On the other hand, RRT* may experience unpredictable performance issues due to its randomize behaviour. Thus, there is no guarantee of the computational cost. In the proposed solution, we have modified original RRT*~\cite{karaman2011sampling} introducing an adaptive search space and a steering function which help to increase consistency of the planner. Besides, we propose multiple RRT* which generate a set of desired paths in which optimal path is chosen. Then apply kinodynamic smoothing, which will result in dynamically feasible as well as obstacle-free path. Thereafter, b-spline based trajectory is generated to maneuver the vehicle autonomously. 

\textbf{Our Contributions: }
\begin{enumerate}
    \item We have modified original RRT* introducing an adaptive search space and a steering function which help to increase the consistency of the planner
    \item Proposing horizon based multiple RRT*. In each of the RRT* instances, it will find the optimal goal location and terminate at newly generated goal pose if the path length exceeds the prediction horizon
    \item Applying iLQR (Iterative Linear Quadratic Regulator) to smooth the optimal obstacle-free path which ensures the feasibility of the dynamic of the vehicle
\end{enumerate}

\section{Related Work}
Reasoning the environment in real-time is important for collision-free planning. Thus, keeping track of the environment (map) close by the current pose of MAV with a predefined perimeter is necessary under which map should be updated in an incremental fashion. 
Hornung. et al.~\cite{hornung2013octomap, steinbrucker2014volumetric} proposed Octomap, which uses probabilistic occupancy estimation to construct the map which uses octree for storing the information. In the process of map building, random measurements (e.g., reflections, dynamics obstacles) should consider. Hence, environment representation with Octomap is concise and accurate enough for planning. Subsequently, voxel hashing~\cite{niessner2013real} is another technique which uses Truncated Sign Distance Fields (TSDFs)~\cite{oleynikova2016voxblox} for reconstructing the environment.  

Path planning can be classified into several levels~\cite{souissi2013path}. Deterministic and probabilistic is one of the ways to classify them. For a given search space, the same result is expected for deterministic approaches, i.e., A* and Dijkstra. Deterministic approaches have several drawbacks including not effectively plan the path satisfying real-time constraints and increase the complexity when the search space dimension is higher, i.e., 3D. In contrast to deterministic approaches, probabilistic based approaches (e.g., Probabilistic Road Mapping~\cite{kavraki1996probabilistic}, Randomly exploring Random Trees~\cite{lavalle1998rapidly}) overcome those constraints despite adding more computation footprint. RRT* (Rapidly-exploring Random Tree)~\cite{karaman2011sampling} is one of the well-adapted path planning techniques which can categorize as a sampling-based technique. 

Most of the trajectory planning algorithms perform path planning followed by feasible trajectory generations as a two-step pipeline;
This may be problematic when it is needed replanning in which path planner unaware of the vehicle’s dynamics. To address this problem, kinodynamic based motion planning~\cite{donald1993kinodynamic} is highly desirable, which ensures the dynamic feasibility. Deterministic path planning algorithms such as A* as well as probabilistic path planning techniques (e.g., RRT*, BIT*~\cite{lan2016bit}, FMT*~\cite{starek2015asymptotically}, etc) can be modified adding kinodynamic capabilities. However, real-time kinodynamic motion planning has been seen as an open problem yet due to high computational cost. 


To generate optimal control for a given system, which is utilized in kinodynamic planning, there are various techniques have been proposed. Linear Quadratic Regulator (LQR) was suggested by Glassman and Tedrake~\cite{glassman2010quadratic} for kinodynamic RRT planner for generating optimal control inputs, provided system dynamics. Alejandro Perez, Robert Platt Jr~\cite{perez2012lqr} extended preceding idea for RRT* while linearizing the non-linear system dynamics at each newly sample point. Moreover, they use infinite horizon LQR policy to obtain optimal control inputs. Instead of using infinite horizon, Jur and Berg~\cite{van2016extended} propose finite horizon iterated LQR and extended LQR smoothing techniques considering non-linear dynamics and non-quadratic cost for optimal kinodynamic motion planning. 

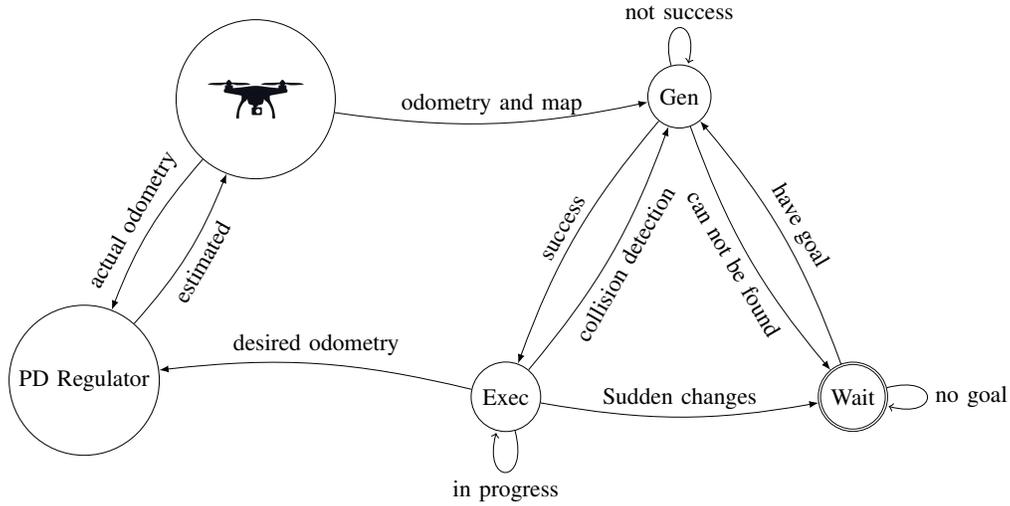
\begin {figure*}
\centering

\resizebox{0.75\linewidth}{!}{
\begin{tikzpicture}[scale=1]
    \node[state] (A) at (90:3) {Gen};
    \node[state] (B) at (210:3) {Exec};
    \node[state, accepting] (C) at (330:3) {Wait};
    \node[state] (D) at (188:9) {PD Regulator};
    \node[state] (E) at (155:7) {\includegraphics[width=.080\textwidth]{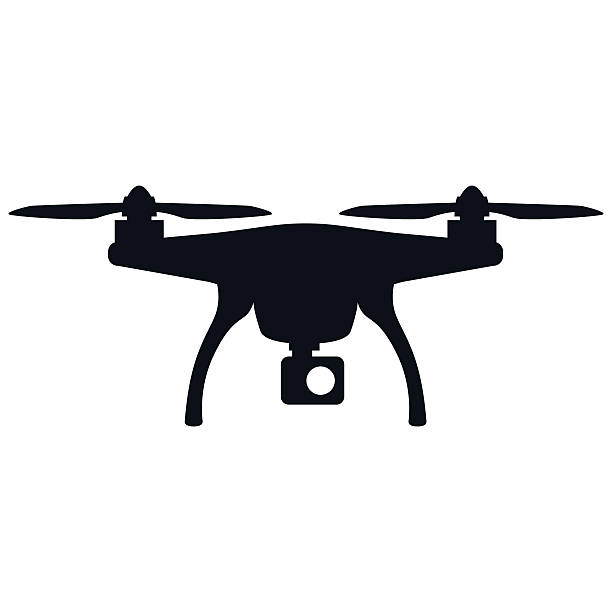}};
    \draw[-latex] (A) to[bend right=10] node[above,rotate=60] {success} (B);
    \draw[-latex] (B) to[bend right=10] node[below,rotate=60] {collision detection} (A);
    \draw[-latex] (A) to[bend right=10] node[below,rotate=300] {can not be found} (C);
    \draw[-latex] (C) to[bend right=10] node[above,rotate=300] {have goal} (A);
    \draw[-latex] (B) to[bend right=10] node[above] {Sudden changes } (C);
    \draw[-latex] (C) edge [loop right] node {no goal} (C);
    \draw[-latex] (B) edge [loop below] node {in progress} (B);
    \draw[-latex] (A) edge [loop above] node {not success} (A);
    \draw[-latex] (B) to[bend right=10] node[above] { desired odometry} (D);
    \draw[-latex] (D) to[bend right=10] node[below,rotate=60] { estimated} (E);
    \draw[-latex] (E) to[bend right=10] node[above,rotate=60] { actual odometry} (D);
    \draw[-latex] (E) to[bend right=10] node[above,rotate=0] { odometry and map} (A);
    \end{tikzpicture}
}
\caption{Initially, the vehicle stays in the wait state until the goal pose is given. If the goal pose within the map and no obstacles are on the goal pose, change its state into Gen which will find the optimal trajectory in between current pose and the goal pose. Afterwards, the state will change into the Exec state until it reaches the goal pose. If there are obstacles close by vehicle, it retries to generate a new trajectory. This process continues until it navigates to the goal pose. Besides, Exec state can be changed into Wait state, if there are some sensor malfunctions which cause the sensor streams interruption. PD (P-proportional, D-derivative) regulator is taking actual and desired odometry and calculate the estimated velocity and yaw angle for controlling the vehicle}

\label{f:overall_idea}
\end{figure*}
Path planning results in a sequence of waypoints which are connected through a set of straight lines and sharpen turns. Thus, a path may not be desirable for navigation due to three constraints: geometric continuity, safety and feasibility of the vehicle dynamics. To fulfil these constraints, the path is to be smoothed ensuring preceding constraints. 
Optimization-based approaches can satisfy all three preceding constraints.  Work in~\cite{zhu2015convex}, proposes a convex elastic smoothing (CES) algorithm, for trajectory smoothing as a convex optimization problem. Timed Elastic Band (TEB Planner)~\cite{rosmann2017kinodynamic} is a kinodynamic planner which locally optimizes the trajectory while considering provided constraints.  Zhou et al.~\cite{zhou2019robust} propose a kinodynamic local planner, based on A* which works quite aggressively. 


B-spline~\cite{usenko2017real}, minimum-snap~\cite{mellinger2011minimum} and its variants are widely used for trajectory generation in the recent past.
B-spline~\cite{babaei2018optimal, biagiotti2010b} are extensively used for trajectory generation due to several reasons. Clear geometrical meaning is one of the main reasons which highly adapted in trajectory generation in 3D space. B-splines, much simpler from the computational point of view because local changes in the trajectory can be done quickly and easily without recomputing the entire trajectory~\cite{bobrow1988optimal}. 

\section{Methodology}
Pictorial visualization of the proposed framework is depicted in Fig.~\ref{f:overall_idea} whereas the workflow is shown in Fig.~\ref{f:workflow}. The following sections explain in detail how each of the components contributes to acceptable system functioning. 

\begin{figure}[!ht]
\begin{center}
\includegraphics[width=1\linewidth]{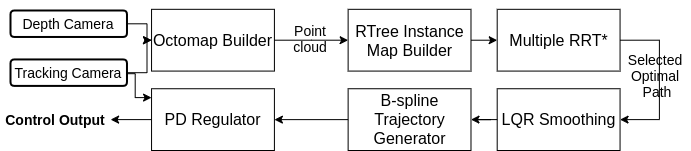}
\caption{The sequence of steps of the proposed framework. Depth camera works at 6Hz whereas tracking camera works at 20Hz. Depth map and odometry constitute the input for the Octomap server which builds the map incrementally. Based on the output of Octomap server, instance map is constructed which is utilized by Multiple RRT* for generating an optimal obstacle-free path. Afterwards, iLQR smoothing followed by trajectory generation is applied on the optimal path. Finally, PD regulator is employed to synchronize the generated and actual trajectories}
\label{f:workflow}
\end{center}
\end{figure}

\subsection{Environment Representation}

Since we use a depth camera for reasoning the environment, constructing incremental map building is necessary because the camera has only (85.2'x58'x94') field of view (FoV) which is not enough for planning. Thus, initially, we feed the camera depth map into Octomap server. Octomap server constructs the map of the environment incrementally. Thereafter, point cloud around the current pose of the vehicle is extracted. RTree~\cite{guttman1984r} is constructed from an extracted point cloud which represents the instance map of the environment.  

\subsection{Adaptive search space}\label{sec:adaptive_search_space}
\begin{figure}[!ht]
\begin{center}
\includegraphics[width=0.9\linewidth]{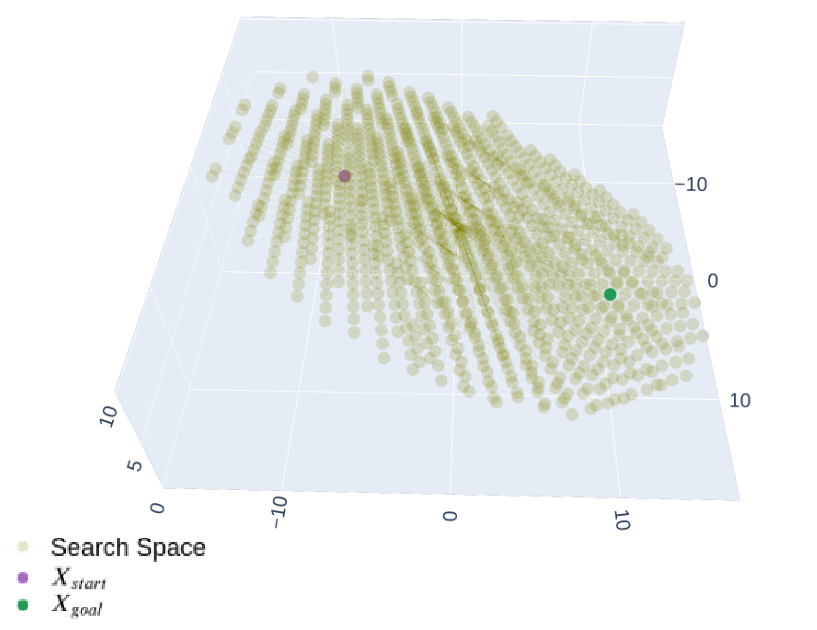}
\caption{Ellipsoidal search space is constructed as we proposed in Algorithm~\ref{alg:search_space} complemented by satisfying the map constraints, i.e., min and max dimension of current search space. This helps to fast convergence of the RRT* planner}
\label{f:precesion}
\end{center}
\end{figure}
RRT* is a well-known technique for path planning in high dimensional spaces, i.e. 3D in which search space defines the whole map of the environment in the default setting. We have made several changes to reduce the execution time and increase consistency. Thus, it is good for global path planning not for local planning specially replanning. On the contrary, in this study, we use RRT* for local replanning. Hence, ensuring consistency of consecutive paths is required. One way to improve consistency is by generating proper random samples. Subsequently, defining the optimal local search space which is closer to the current pose of the vehicle. Thus, we proposed a deterministic way of generating search space which eventually helps to improve the consistency of the planner significantly. 


Search space is being constructed as an ellipsoidal search space (spherical or oblate spheroid or prolate spheroid) in which principal axes are defined by start and goal pose of the trajectory. In general, random samples can be generated within constructed ellipsoid, but we proposed a deterministic way of generating sample points (Algorithm~\ref{alg:search_space}) as similar to~\cite{ellipsoid} while considering constraints of the traversable space. Besides, this helps to reduce the execution time. Since search space is being changed according to the traversable space around the current pose of the vehicle, we call it as an adaptive search space. RRT* planner picks points randomly from the deterministic search space. 

Adaptive search space is defined as 
\begin{equation}
    (x-c_x)^2/(r_x)^2 + (y-c_y)^2/(r_y)^2 + (z-c_z)^2/(r_z)^2 = 1
\end{equation}
where c is the middle pose in between start and goal pose of the trajectory and $\mathbf{r} = <r_x, r_y, r_z>$ is defined as follows:
\begin{equation}
\begin{split}
    \mathbf{r} = \mathbf{x_{goal}} -\mathbf{x_{start}}
\end{split}
\end{equation} where  $\mathbf{r}_x = max(4.0,  \mathbf{r}_x), \; \mathbf{r}_y = max(4.0,  \mathbf{r}_y) \;$ and $ \mathbf{r}_z = max(4.0,  \mathbf{r}_z)$. The rotation matrix R is calculated between $\mathbf{z}$ (0,0,1) and $\mathbf{r}$ as given here~\cite{kulathunga2020realtime}. 

\begin{algorithm}[h!]
\SetAlgoLined
\textbf{input data} : 
$n$ points to be generated, 
$r$ principal semi-axes, R rotation matrix
$c$ center position \\
$cols = 0, \; n_p = 0, \; r_{min} = min(r), \;M = zero(3, n)$ \\

\If{$r_{min} == r_x$}{ 
        $h = 2*\frac{r_x}{2n+1}$,
        $n_i = n$,
        $n_j = \frac{r_y}{r_x}*n$,
        $n_k = \frac{r_z}{r_x}*n$ 
} \If{$r_{min} == r_y$}{ 
        $h = 2*\frac{r_y}{2n+1}$,
        $n_j = n$,
        $n_i = \frac{r_x}{r_y}*n$,
        $n_k = \frac{r_z}{r_x}*n$ 
}
\Else{
        $h = 2*\frac{r_z}{2n+1}$,
        $n_k = n$,
        $n_i = \frac{r_x}{r_z}*n$,
        $n_j = \frac{r_y}{r_z}*n$ 
}
\For{$k = {0,...,n_k}$}{ 
{}{
    $z = c_z + k*h$\\
    \For{$j = {0,...,n_j}$}{ 
    {}{
        $y = c_y + j*h$\\
        \For{$i = {0,...,n_i}$}{ 
        {}{
            $x = c_x + i*h$\\
            $M(0) = <x, y, z>$ \\
            
            \For{$l = {0,1,2,...,7}$}{ 
            {}{
                \If{$l == 0 || l == 1 || l == 4$}{ 
                {}{
                    $n_p = 0$
                }
                }
                \If{$l \leq	 0 $}{ 
                {}{
                    $M(cols+l) = <2c_x - M(n_p)_x, M(n_p)_y, M(n_p)_z>*R$
                }
                }
                \If{$ 1 \leq l \leq	 2 $}{ 
                {}{
                    $M(cols+l) = <M(n_p)_x, 2c_y - M(n_p)_y, M(n_p)_z>*R$
                }
                }
                \If{$l \geq	 3 $}{ 
                {}{
                    $M(cols+l) = <M(n_p)_x, M(n_p)_y, 2c_z - M(n_p)_z>*R$
                }
                }
                $n_p++$\\
                $cols+=8$
            }
            }
        }}
    }}
}}
\caption{Generating points over the interior of the search space}
\label{alg:search_space}
\end{algorithm}

\subsection{Multiple RRT*}
\begin{figure}[!ht]
\begin{center}
\includegraphics[width=0.9\linewidth]{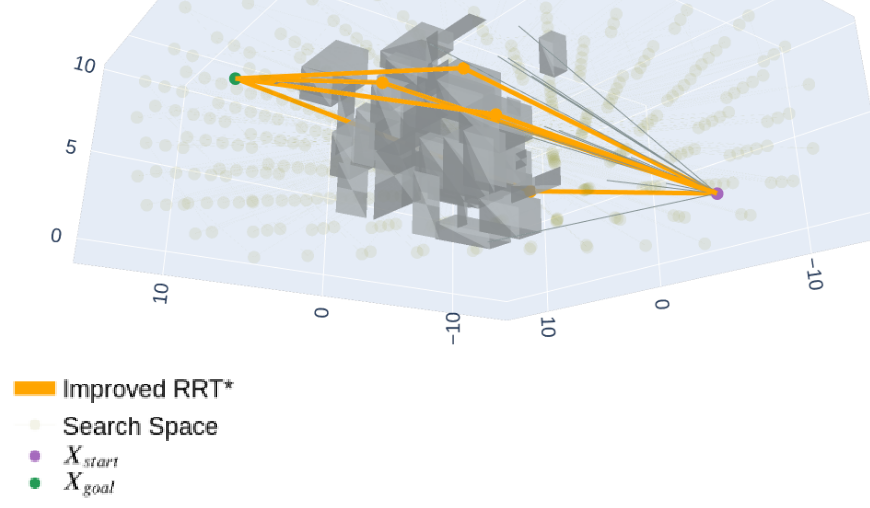}
\caption{Multiple RRT* can generate a given number of paths in between start and goal pose (i.e., four paths) in this scenario}
\label{f:multi_rrt}
\end{center}
\end{figure}
To further improve the consistency, we use multiple RRT* instances using a thread pool which utilizes multi-cores/multi-processors will result in generating N number of paths. N is a configurable parameter. It is better to select N as the number of cores in your embedded computer because the thread pool is created in the phase of algorithm initialization with numbers of threads which equals to the number of system cores, which helps to improve performance. Initially, select the path that has the lowest cost as the optimal path. Cost is given by:
\begin{equation}
    Cost = \left\Vert\left(\bf P_{\rm M} - \bf P_{\rm {goal}}\rm\right) \right\Vert  + \sum_{m=1}^{M-1} \left\Vert\left(\bf P_{\rm m} - \bf P_{\rm {m+1}}\rm\right) \right\Vert  
\end{equation} where $\bf P_{\rm {m}}$ depicts the mth waypoint of the selected path in which it consists of M number of waypoints. Always, it is not correct to consider the lowest cost belongs to the optimal path. We have to check the safely of the path. If the chosen path closer to obstacles, next time it will pick the path which has the second-lowest-cost. If the environment is cluttered, the optimal selection will go up to the highest cost as well. Once an optimal path is selected and start moving on the trajectory, next time, it will start from the lowest cost. If the distance between start and goal pose higher than a predefined value (planning horizon), RRT* will return path up to the planning horizon; This is the procedure for selecting an optimal path from generated paths.

\subsection{Path smoothing via iLQR}
An optimal path which chooses from the preceding step is to be smoothed while considering obstacles around the vehicle. The optimal path consists of a set of waypoints which connect start and goal pose. Next step is to take consecutive three waypoints and get midpoints of first and second waypoints and second and third waypoints and apply iLQR between those two midpoints. iLQR is solved as a finite horizon (N steps) optimization problem. N, the number of steps at max is required to emulate vehicle dynamics.

We use the same quadcopter model as given in ~\cite{van2014iterated}. Let the system state space be $x = [p^T v^T r^T w^T]$  where p,v,r and w stand for position (m), velocity (m/s), orientation about axis $\mathbf{r}$ by angle $|r|$ (rad) and angular velocity (rad/s) respectively. Size of the state space ($x$) equals to 12 (n) and system has 4(m) control inputs : $u_1, u_2, u_3$ and $u_4$. Vehicle continuous-time dynamics $\dot{x} = f(x,u)$ is given as follows:
\begin{equation}
\begin{split}
       & \dot{p} = v \\
       & \dot{v} = -ge_3 + ((u_1-u_2+u_3+u_4)exp([r])*e_3-k_vv)/m \\
       & \dot{r} = w + \frac{1}{2}[r]w+\frac{(1-\frac{|r|}{2tan(\frac{1}{2}|r|)})[r]^2w}{|r|^2} \\
       & \dot{w} = J^{-1}(\rho(u_2-u_4)e_1 + \rho(u_3-u_1)e_2 \\ & \; \; \;\;\;\;\; + k_m (u_1 - u_2 +u_3-u_4)e_3 -[w])Jw)
\end{split}
\end{equation}
where $e_i; i=1,2,3$ standard the basis, g stands for gravity, $k_v$  and $k_m$ are constants, m, J, $\rho$ are mass (kg), moment of inertia (kg$m^2$) and distance from center of vehicle to center of rotor (m) respectively. $[a]$ depicts the skew-symmetric matrix notation. Afterwards, it is needed to define an optimal control policy ($\pi \in \mathbb{X} \rightarrow \mathbb{U}$) where $\mathbb{X} \subset \mathbb{R}^n$ and $\mathbb{U} \subset \mathbb{R}^m$ utilizes for generating a path in-between given two poses ensuring the vehicle dynamics and environmental constraints. Since system dynamics is non-linear at each considered discrete time instance, system dynamics is to be linearized as follows:

\begin{equation}
    \begin{aligned}
     f(x_k, u_k)\approx \tilde{f}(x_k, u_k) = \\ f(\bar{x_k}, \bar{u_k}) + \frac{\partial f}{\partial x}(\bar{x_k}, \bar{u_k})(x_k-\bar{x_k})  + \frac{\partial f}{\partial u}(\bar{x_k}, \bar{u_k})(u_k-\bar{u_k})
    \end{aligned}
\end{equation} where $\bar{x_k}$ and $\bar{u_k}$ are the estimated values at kth step. $\bar{u_k}$ is estimated with the help of Riccati equation~\cite{marro2002geometric}. Then the optimal control inputs ($u_0^*, u_1^*, ..., u_{N-1}^*$) can be calculated as follows:

\begin{mini!}|l|[2]
{u_0,...,u_{N-1}}{\sum_{k=0}^{N}{c(x_k, u_k)}}
{}{}
\addConstraint{x_1 = f(x_0, u_0)}
\addConstraint{x_2 = f(x_1, u_1)}
\addConstraint{...}
\addConstraint{x_N = f(x_{N-1}, u_{N-1})}
\end{mini!} where $x_0$ is a starting pose where the path is to be smoothed. Linearized cost $c(x_k, u_k) \approx \tilde{c}(\delta x_k, \delta u_k)$ can be defined as:
\begin{equation}
    \begin{aligned}
      \tilde{c}(\delta x_k, \delta u_k) =  c(\bar{x}_k, \bar{u}_k) + \Delta c(x_k, u_k)\begin{bmatrix}
 x_k - \bar{x_k} \\
 u_k - \bar{u_k}
\end{bmatrix}+ \\ \frac{1}{2} \begin{bmatrix}
 x_k - \bar{x_k} \\
 u_k - \bar{u_k}
\end{bmatrix}^T\Delta^2 c(x_k, u_k)\begin{bmatrix}
 x_k - \bar{x_k} \\
 u_k - \bar{u_k}
\end{bmatrix} + q\Sigma_iexp(-d_i(x_k))
    \end{aligned}
\end{equation} where q is a scaling factor and $d_i(x_k)$ is the sign distance between ith obstacle and considered pose $x_k$. Once the optimal control inputs are obtained by applying 4th order Runge-Kutta integrator, projected poses can be calculated. Those poses represent the smoothed path from $x_0$ to intermediate goal pose.

\begin{figure}[!ht]
\begin{center}
\includegraphics[width=0.85\linewidth]{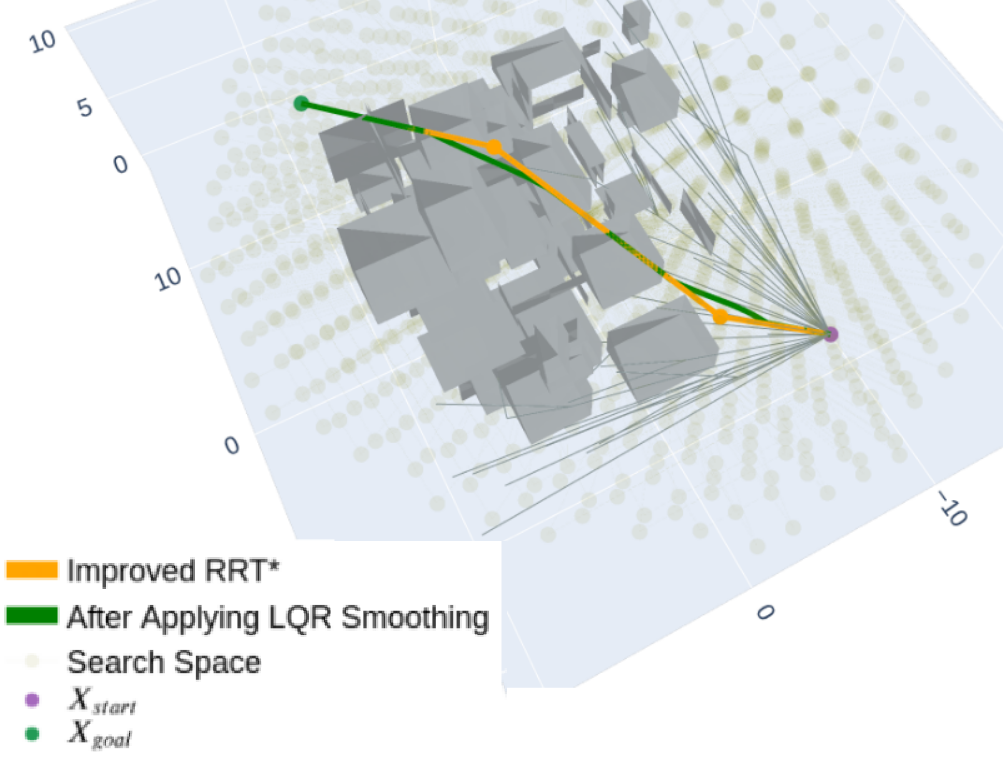}
\caption{The proposed multiple RRT* is capable of selecting an optimal path out of the generated paths. Afterwards, the optimal path is smoothed by LQR smoothing which depicts in green color}
\label{f:rrt_lqr}
\end{center}
\end{figure}

\section{Experimental Results}
\begin{figure}[!ht]
  \begin{subfigure}[b]{1\linewidth}
    \centering
    \includegraphics[width=0.85\linewidth]{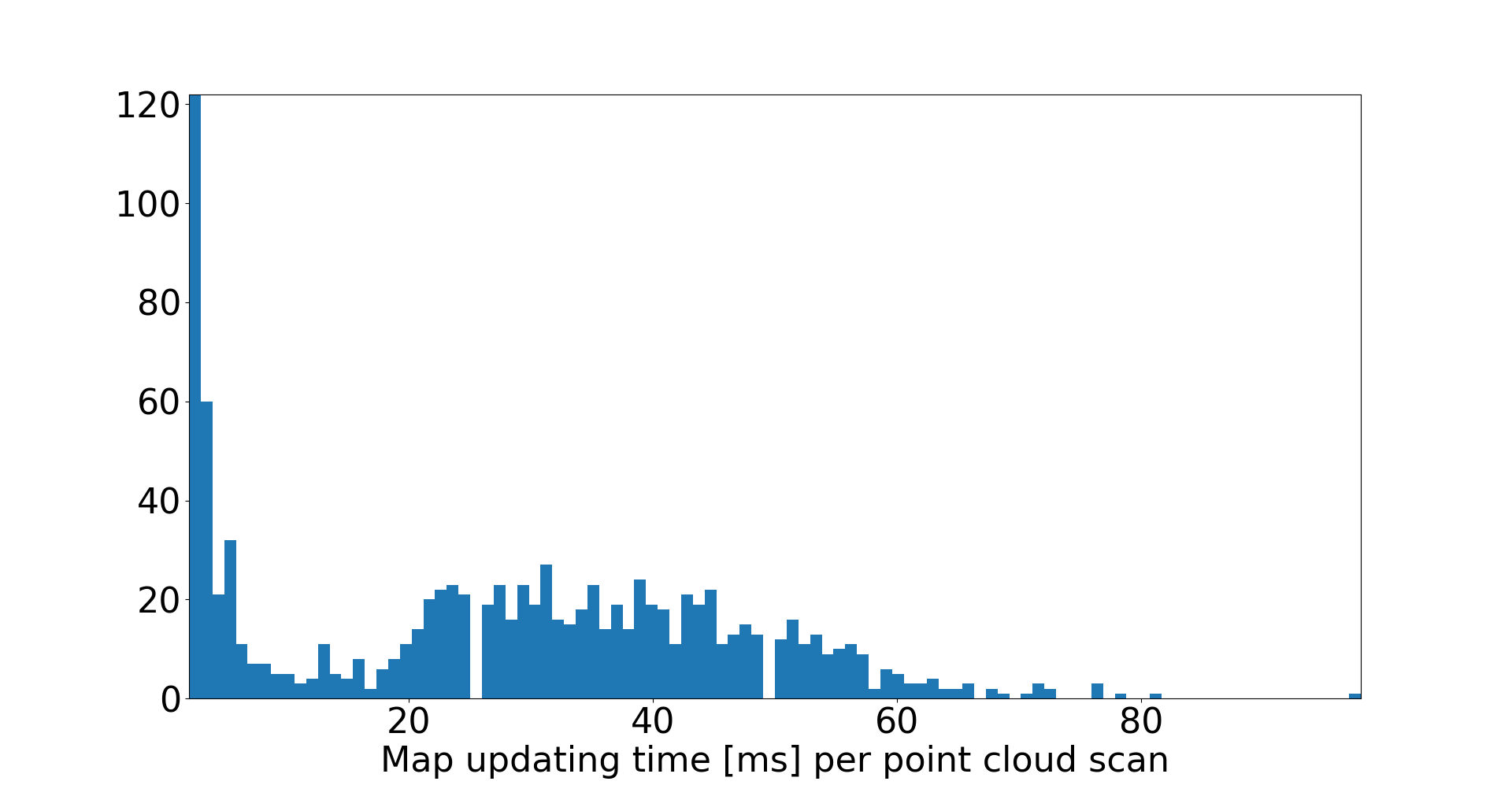}
    \caption{3D Ring Buffer~\cite{usenko2017real}}
    \label{fig:map_ring_buffer}
  \end{subfigure}%
  \hfill
   \begin{subfigure}[b]{1\linewidth}
    \centering
    \includegraphics[width=0.85\linewidth]{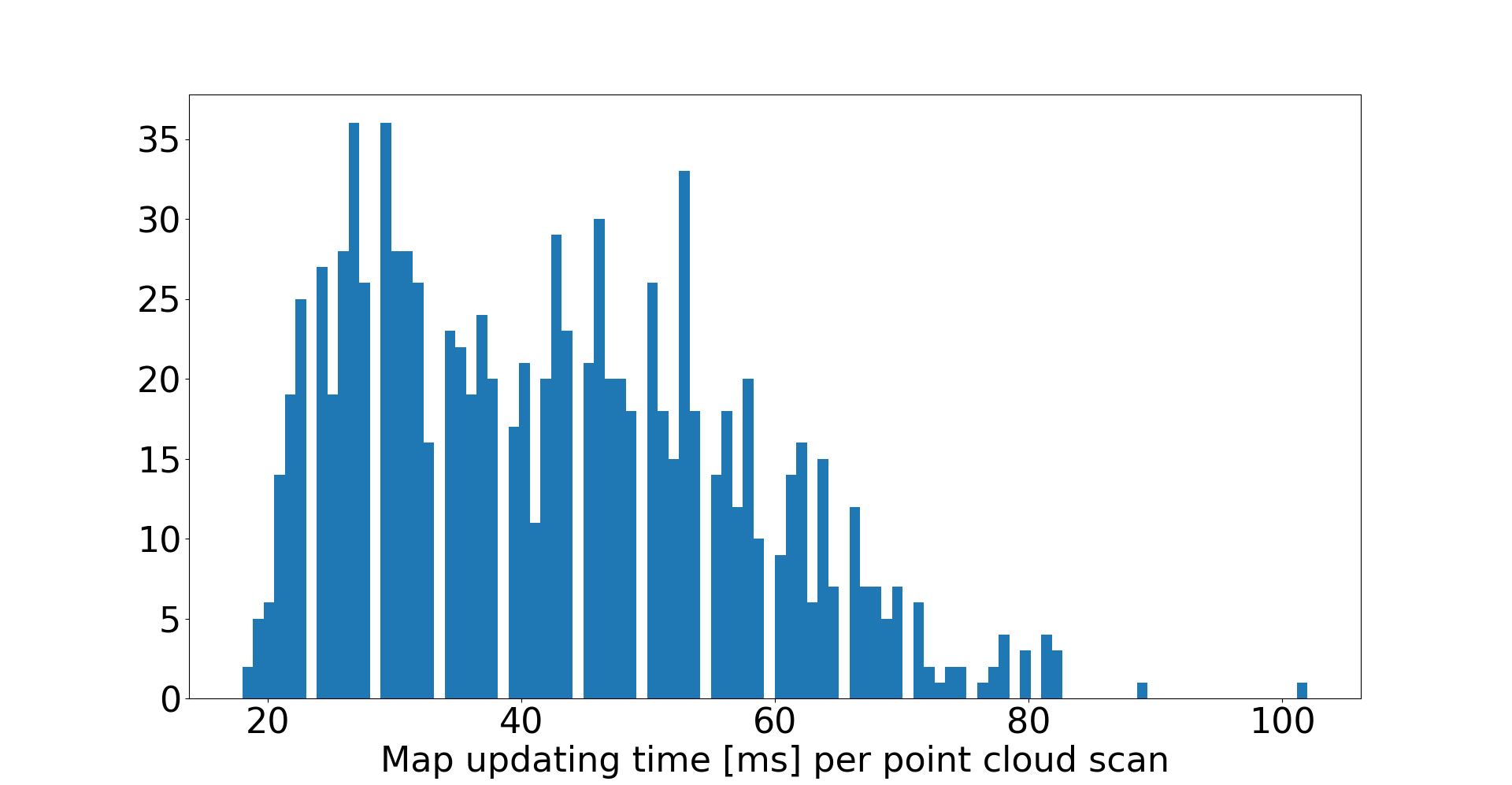}
    \caption{Euclidean Signed
Distance Field~\cite{zhou2019robust}}
     \label{fig:map_edt}
  \end{subfigure}%
  \hfill
  \begin{subfigure}[b]{1\linewidth}
    \centering
    \includegraphics[width=0.95\linewidth]{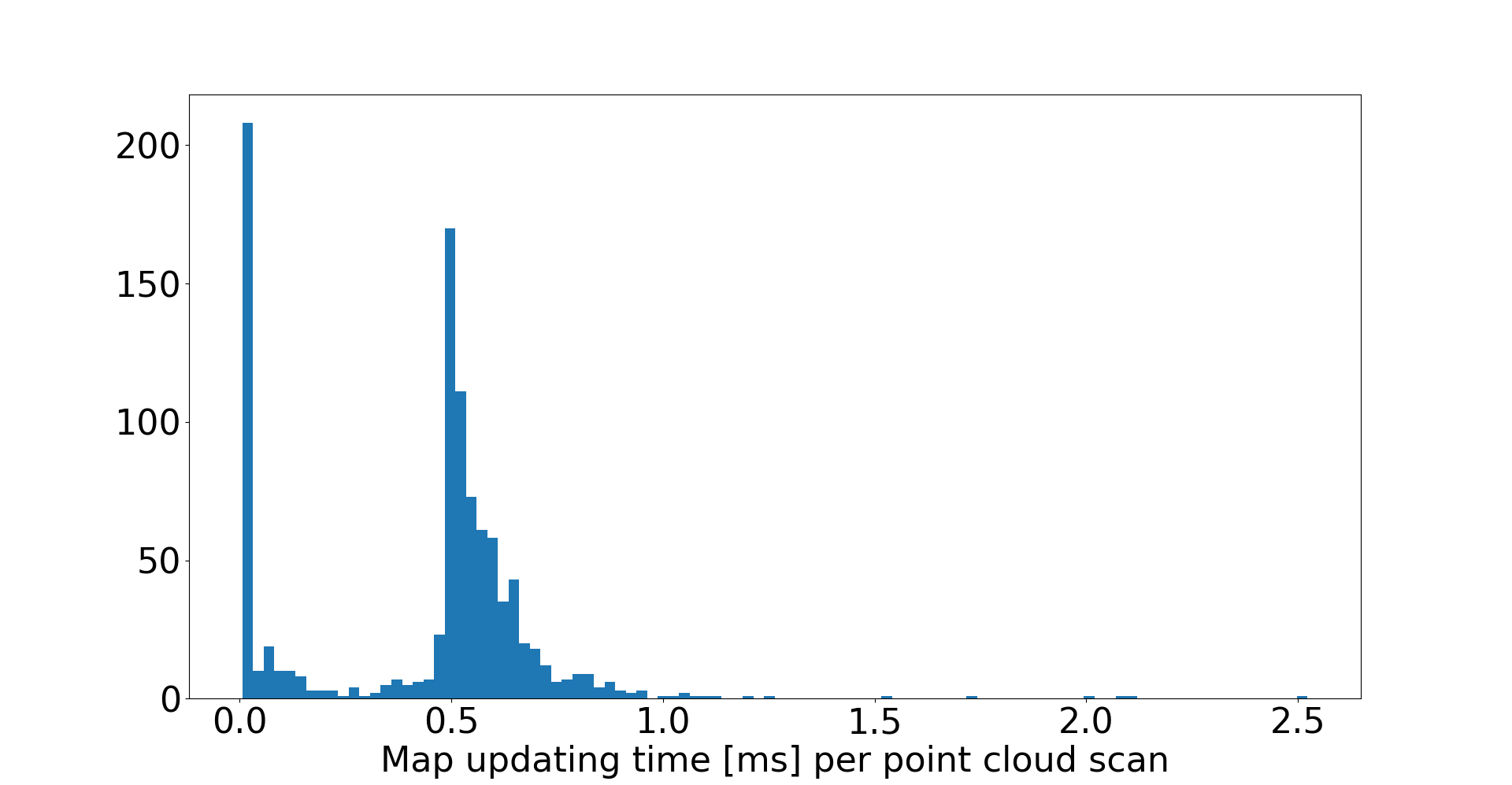}
    \caption{Proposed instance map building approach}
     \label{f:map_rtree}
  \end{subfigure}
  \caption[Obs Point 1]{Comparison of map updating time (ms) per point cloud scan. Range of the map set for 4m around the quadcopter's current pose. Subplots (a), (b), (c) show the updating time as histogram}
  \label{f:map_building}
\end{figure}

In this section, we present qualitative and quantitative validation of the proposed local replanner. First, we validate our proposed instance map building with two other techniques. Afterwards, the performance of the improved RRT* is compared with A*  and original RRT*. Then, the complete system is evaluated in various simulated environments. The proposed planner is implemented in C++11 in that the sparse matrix library Eigen is used while enabling GNU C++ compiler optimization level to -O2. All the simulated experiments are conducted on a computer (Intel(R) Core(TM) i7-7500U CPU @ 2.70GHz CPU and 8 GB RAM).

RTree~\cite{guttman1984r} based instance map is constructed when the depth map is available at 6Hz. We have adopted this approach to reduce the execution time, which utilizes for map building on the quadcopter. We have evaluated our instance map building approach with two other existing approaches as shown in Fig.~\ref{f:map_building}. In the proposed approach, it is needed to keep two instance maps (current and previous). The previous map is utilized by the planner. Once the current map is built, the previous map replaces with the current map. Despite building an incremental map (~\cite{usenko2017real, zhou2019robust}), the proposed approach will save computational cost and execution time while maintaining accuracy as similar to those incremental map building approaches.



We have compared the mean computation time of the improved RRT* with original RRT* and A*. All three algorithm are utilized the same search space that consists of 50 random obstacles within a cube of 20 m. We have generated 100 different search spaces and calculated mean computation time. Result is given in Table.~\ref{ta::mean_time_rrt_and_improved_rrt}. In Fig.~\ref{f:rrt_with_imporved}, it is shown an example scenario. We were able to reduce the computation time considerably. Besides, improved RRT* has a constrained search space, it helps to improve the consistency of the path finding as we explained in the section~\ref{sec:adaptive_search_space}. 
Also, we calculated the mean computation time of the main five operations that involve in the proposed planner as percentage values as shown in Fig.~\ref{f:execution_time_mean}. 

\begin{table}[ht!]
\caption{Mean computation time for path finding under which same search space utilized for 100 different trials}
\label{ta::mean_time_rrt_and_improved_rrt}
\centering
\begin{tabular}{|l|l|l|l|}
\hline
& A*     & Original RRT* & Improved RRT* \\ \hline
\begin{tabular}[c]{@{}l@{}}Mean Computation\\ time (s)\end{tabular} & 1.9612 & 1.2512  & 0.9621        \\ \hline
\end{tabular}
\end{table}

\begin{figure}[!ht]
\begin{center}
\includegraphics[width=0.85\linewidth]{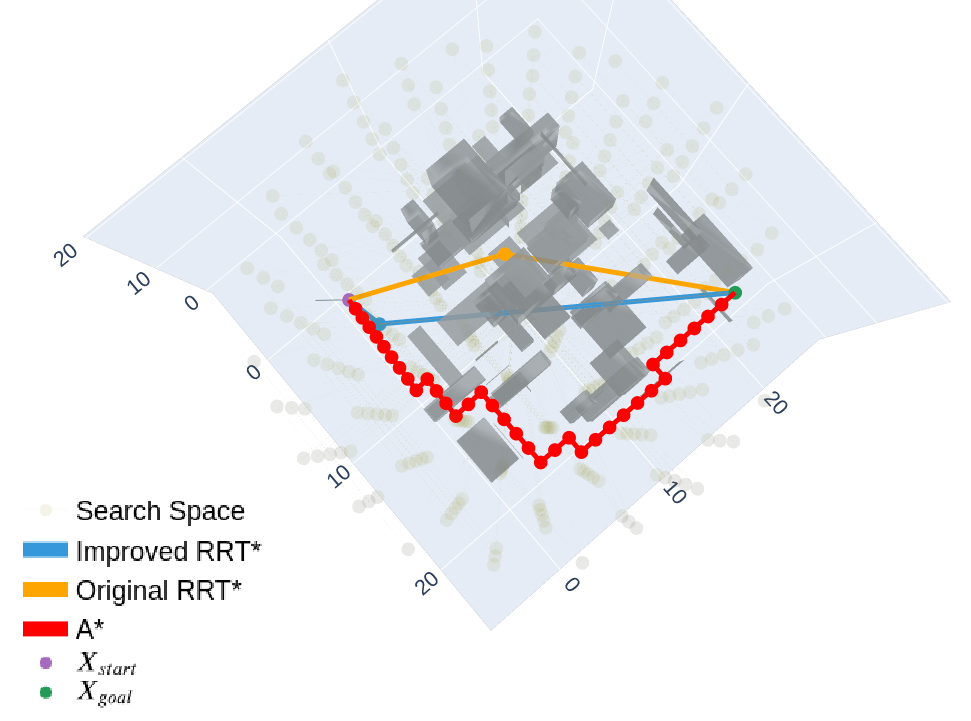}
\caption{Example scenario of evaluation we performed (Table.~\ref{ta::mean_time_rrt_and_improved_rrt}) to estimate average execution time of improved RRT* compared to original RRT* and A*}
\label{f:rrt_with_imporved}
\end{center}
\end{figure}

\begin{figure}[!ht]
\begin{center}
\includegraphics[width=0.9\linewidth]{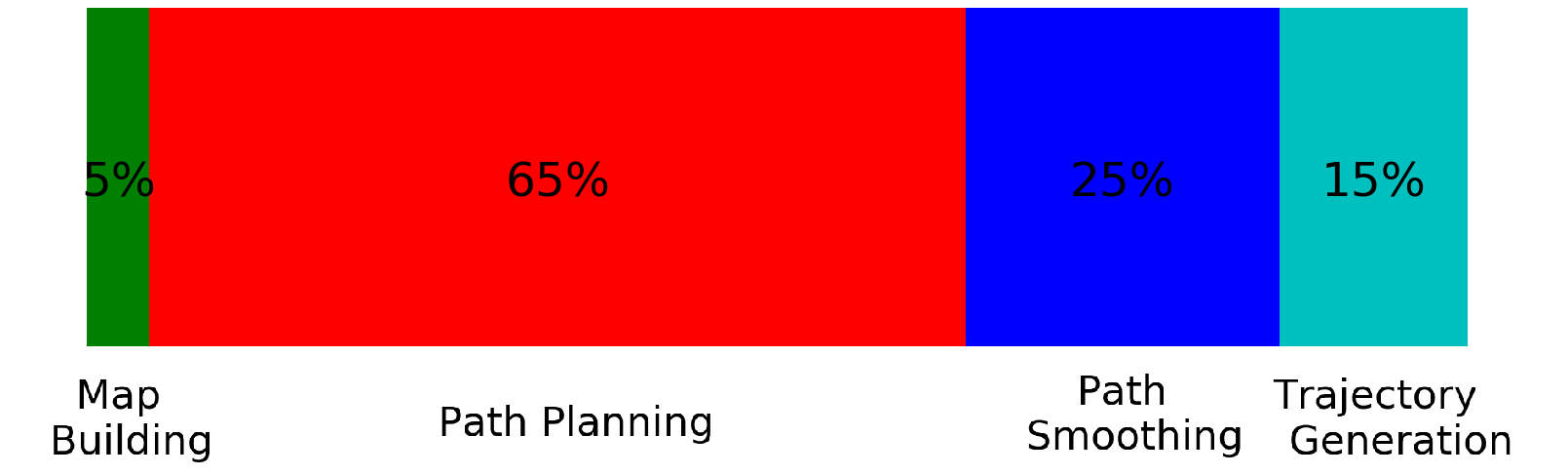}
\caption{Mean computation times of the main four operations that involved in the proposed planner as percentages}
\label{f:execution_time_mean}
\end{center}
\end{figure}

Finally, we have conducted a few experiments on different simulated environments. In Fig.~\ref{fig:simulated_environment}, one of the selected simulated environment for testing the behaviour of the planner. When the distance between goal and start position less than the predefined horizon (Fig.~\ref{fig:full_path}), it generates complete path or else it generates path up to the horizon as shown in Fig.~\ref{fig:horizon}.

\section{Acknowledgment}
The work presented in the paper has been supported by Innopolis University, Ministry of Education and National Technological Initiative in the frame of creation Center for Technologies in Robotics and Mechatronics Components (ISC 0000000007518P240002)

\section{Conclusion}
We presented a framework for path planning which followed by kinodynamic smoothing while ensuring the vehicle dynamics feasibility for MAVs. We have chosen "RRT*" for path planning in which several improvements have done to reduce the planning time, which helps to employ as a local planner. Kinodynamic smoothing ensures the dynamic feasibility and obstacle-free path. Finally, we have validated the proposed framework in various simulated environments to check the behaviour of the planner. Our validation result shows the performance of the planner. In future works, we are going to improve the efficiency of the path planner. 

\bibliographystyle{IEEEtran}

\bibliography{references}

\end{document}